# Real-time pothole detection with onboard sensors and camera on vehicles

Aswath Muthuselvam, Jeevak Raj S, Mohanaprasad K

School of Electronics Engineering, Vellore Institute of Technology, Chennai
{m.aswath2017,sjeevak.raj2017}@vitalum.ac.in, kmohanaprasad@vit.ac.in

**Abstract.** Road conditions play an important role in our everyday commute. With the proliferating number of vehicles on the road each year, it has become necessary to access the road conditions very frequently, this would ensure that the traffic also flows smoothly. Even the smallest crack in the road could be easily be chipped into a large pothole due to changing surface temperatures of the road and from the force of vehicles riding over it. In this paper, we have addressed how we could better identify these potholes in realtime with the help of onboard sensors in vehicles so that the data could be useful for analysis and better management of potholes on a large scale.

For the implementation, we used an SVM classifier to detect potholes, we achieved 98.1% accuracy based on data collected from a local road for about 2 km which had 26 potholes distributed along the road.

Code is available at: https://github.com/aswathselvam/Potholes

## 1  Introduction

Real-time pothole detection is needed to report any potholes that are taking root in the roadways. Informing authorities of the potholes' location and its effect on vehicles, By reporting to the authorities, it becomes much easier to plan, manage and fix the potholes. This ensures safety for drivers, prevents accidents that occur due to driving over potholes, Vehicle damage in wheel rims which require the owners to change the whole alloy rim which is expensive and also non-economical. Furthermore, the collected information about the potholes can be used for Alerting drivers of upcoming potholes in their route. The digital platform for this system has easier manageability and precise outlook. Massive data collection is possible if implemented on large scale. Based on the data, it is also easy to estimate the cost required for fixing the potholes, the typical cost to repair potholes is approximately $35 to $50 per pothole. There may be an initial mobilization cost of about $100 to $150 [1] to bring trucks and crew out to the repair site.

## 2  Literature review

Hsiu-Wen Wang [2] used a combination of acceleration thresholds, Z-Thresh, Z-DEV, STDEV Z, for detecting potholes. If readings cross the threshold, a pothole is detected. Disadvantages of this method can be false triggers or failure to detect different types of potholes while driving at different speeds and different terrain. This gave us an overview of the existing algorithms' performances.

Umang Bhatt et al. [3] used SVM model to classify the sensor data. They used a total of 26 features formed with raw data collected from sensors. This might require more computational time for the CPU. We try to address this by experimenting with a lesser number of features.

Artis Mednis et al. [4] evaluated various methods like Z-THRESH, Z-DIFF, STD-DEV(Z), G-ZERO algorithms on a mobile running Android Operating system. Each detection method is evaluated with a range of values of threshold points in terms of ratio of G-force. The percentage of correctly detected potholes to the original pothole counts are shown for each methods. Ability of correctly detecting a various types of potholes such as large, small, pothole clusters, gaps and drain pits were also evaluated.

Singh, J., Shekhar, S [5] used Nvidia GPU to detect potholes and this provided us with the insight to explore such options and their choice of the neural net was also explained.

Chen Hanshen, Yao Minghai, Gu Qinlong, [6] explored a different approach to localize and detect a pothole on the road instead of searching for it in the whole image, but rather localizing certain regions and then continue the search.

## 3  Methodology & implementation

We explore 3 modes of pothole detection as follows:

1. Offline method: the user captures the image and uploads it to the server.
2. Online method: the user starts a live stream to detect pothole in realtime.
3. Online method: using accelerometer and gyroscope to detect potholes in realtime.

The computer vision algorithm can detect the presence of pothole from the image both at day and night time and can extract its intrinsic parameters such as dimension( length, width), area and the effect of riding. If the pothole is detected, the app acquires the latitude and longitude position of the device at that instant, and captures the image of the pothole and sends it to the SQL database server, which is running on Node-RED platform. Once the pothole is detected and sent to the server the image of the pothole can be viewed in the maps of Node-RED server. If the user feels that taking photos of the pothole is difficult or if there are many potholes present then the live stream option can be used. The potholes detected from the live stream are uploaded to the server automatically. During night time or in heavy traffic, taking photos is difficult and here comes the use of the third option, with the help of IMU(Inertial Measurement Unit).

For inertial motion sensors, in case of roads with heavy traffic, the camera can't see the road because of dynamic obstacles like other vehicles on the road. This is where the Accelerometer and Gyroscope sensors is a useful substitute. The overall system architecture is shown in Figure 1

### 3.1  Image processing from the captured image

The aim is to detect the pothole exactly and apply a mask to it. To do this we are using the algorithm called Mask RCNN which was developed by FAIR which uses the concept

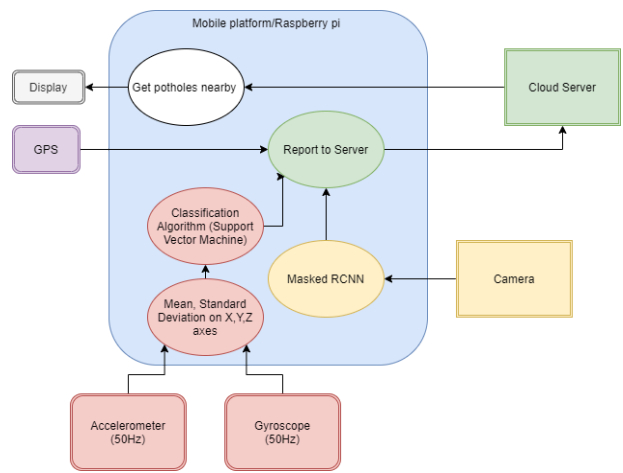

**Fig. 1.** System Architecture

of instance segmentation and Object detection. The object detection technique is used to classify the object and label it with a confidence score. Instance segmentation is used to separate things into groups, isolating the pixels associated with each object

**Training** First, the dataset was collected, which consisted of 1500 images of potholes each with different camera perspectives and under different lighting conditions, even during day and night time, wet and dry scenarios. The images were separated into two classes, train and validation, in a ratio of 90% for training and 10% for validation. For labelling the region of interest, the open-source VIA tool, developed by Oxford University, was used to draw a polygon around the potholes to label it. Then the JSON file with the ground truth was used for training the model for 160 epochs and the model was tested to provide an accuracy of 81%.

**Testing** Once the model has been trained, it was tested in real-time. Good and expected results both at day and night time was obtained. To improve the visibility at night time, logarithmic contrasting and stretching algorithm was employed. As a result, the potholes are highlighted and the area is also obtained. Based on the area, the effect of riding is calculated. The area and effect parameters are added into the list along with the latitude and longitude values, we used free stack API to create the URL of the image and added that to list and sent to the server through socket.

**Measures to prevent duplicate entries** There is a possibility of detecting the same pothole multiple times on different days or by different vehicles. In such cases, the same potholes will be reported multiple times. Marking and localizing the pothole with respect to the surrounding may not be feasible, i.e. surroundings keep changing. So we devised an algorithm that receives nearby potholes data and stores it on the edge device, next it compares the length, breadth, or area of detected potholes with the images obtained from the cloud, if the difference is significant then it uploads to the server as a record of newly detected pothole.

3.2 **Pothole detection in live stream** Detection through live-stream needs a faster algorithm to capture the details instantaneously so we preferred the canny edge algorithm which is quite faster than the Mask-RCNN so that the pothole is detected along with its count and area values and updated to the server.

**Canny Edge detection** It is an edge detection technique used to detect edges in an image, we tuned the edge detector so that the edges of the potholes are highlighted and the value of the threshold came around 0.6, next we dilated the image into a binary image taking a kernel of ones.

**Finding the potholes and its area** The Mask-RCNN gives an image with highlighted pixel mask for region with potholes. The area of the potholes is calculated as the ratio of number pixels in the highlighted region to the total number of pixels in the image.

**Uniformity** Suppose the potholes are detected from a distant view, then the area would be small and when compared to detecting it from a closer position. So to avoid this discrepancy and maintain uniformity, a horizontal threshold line in the image was taken, and only when the pothole overlaps with the virtual line, the detected pothole is registered. As an alternative solution, homography can also be used to get a birds eye view of the dashboard camera feed, to obtain the correct area, irrespective of the distance.

### 3.2 Accelerometer and Gyroscope

For the classification of sensor data as potholes or plain road, a Machine learning classifier called SVM(Support Vector Machines) was used. For real-time implementation, a smart-phone centric mechanism was explored. Two Apps were built -one for data acquisition and another for real-time prediction. The Data acquisition app was built with the MIT App Inventor V2. Then the SVM Machine learning classifier was trained with the collected data. For the SVM classifier, six features were used for the model, that is, mean and standard deviation of Accelerometer and Gyroscope. The support vectors are then used to classify data as either potholes or plain road. The advantages of is approach are that it is more accurate and reliable than thresholding.

**Data Acquisition** Data collection was decided to be done with the help of smartphone sensors. Input data was formatted into seven fields as follows: Time-stamp, Accelerometer X, Accelerometer y, Accelerometer Z, Gyroscope X, Gyroscope y, Gyroscope z, Pothole- detected flag. These are collected at 50Hz. The user opens the app as shown in Figure 2 and starts logging accelerometer and gyroscope sensor data. The user presses the 'pothole' button every time the vehicle drives over a pothole. This associates the acceleration reading for the subsequent 1 second period as a pothole data. The phone was fixed to the floorboard of the vehicle, and gently the 'Pothole' button was pressed when the driver rides the vehicle over the potholes. Around 300Kb of data was collected for training.

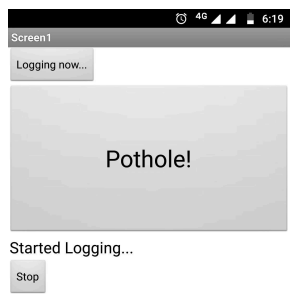
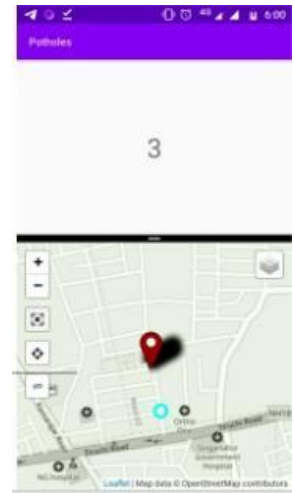

**Fig. 2.** Data Acquisition App                              **Fig. 3.** Detection App

**Data classifier creation** Classification of potholes from the images are done with Mask-RCNN and it was trained on Google Colab. As a quick alternative prototype method, Canny edge detection was also tested in MATLAB, however, there were a lot of false triggers. Mask-RCNN makes pixel classification possible, this way, the exact area of the pothole is calculated. The Mask-RCNN model is also robust to various lighting conditions, it has been tested for both day and night detection as well. Problems addressed with the video feed from the camera are area calculation, detection during both day and night, and the effect of riding on a pothole.

The accelerometer and gyroscope data obtained are visualized in a 3D graph plot for a better understanding and to identify which classifier should be used. As seen in Figure 4 and Figure 5, a linear classifier can be used to classify the pothole and non-pothole data.

The SVM model was prototyped in MATLAB with linear kernel for two-class learning. A test bench was created which calls the entry-point functions and passes the features as parameters. After testing and validation, the entry-point function was exported to C++ code for the ARM Cortex-A processor using MATLAB Coder. The generated C++ code is copied into the Android Studio Projects folder under the JNI directory. Then JNI interface was used to call C++ SVM predict entry-point function by passing the six features as input. On the Android device, the minimum, maximum, mean, standard deviation of the accelerometer and the gyroscope separately were calculated, totaling up to 6 features, and feeding it as input to the SVM classifier. If a pothole is detected, 1 second duration was given to refresh the sensor data values stored in the array which is used to calculate the mean and variance, during this period, the values are not passed into the SVM Classifier.

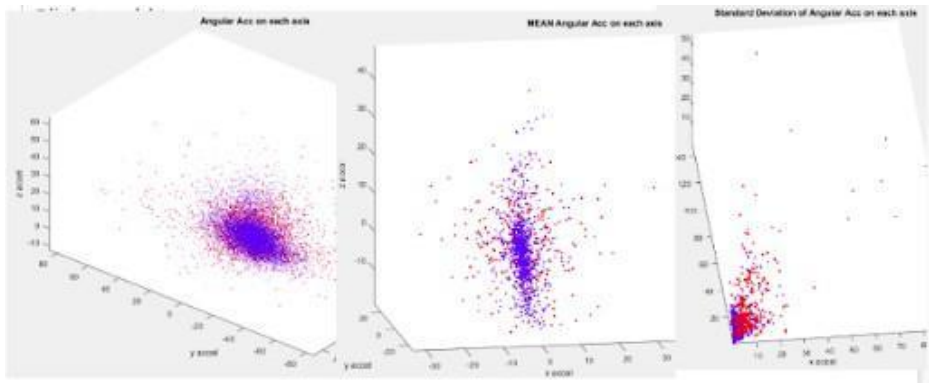

**Fig. 4.** Accelerometer data points

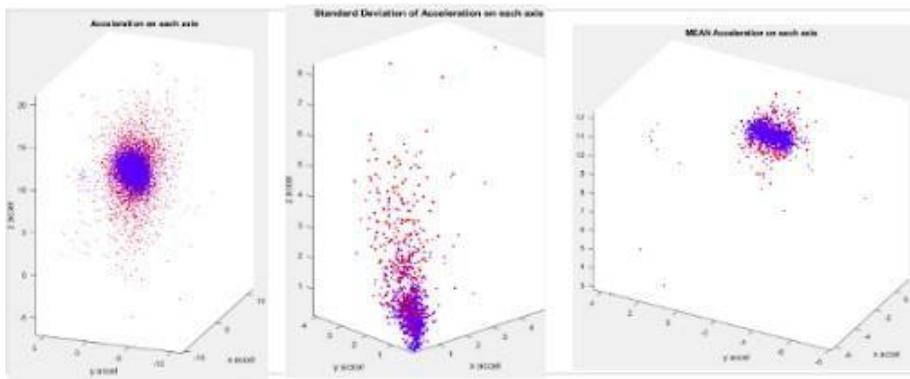

**Fig. 5.** Gyroscope data points

This is to prevent multiple triggers of the same pothole, as new data points obtained just after the detection would also be most likely classified as a pothole. Problems addressed in this section is the detection of a pothole when the camera is not available or the road is not in view, possibly due to heavy traffic. In these cases, the location of the pothole and the effect of riding on it is reported.

The code for the Detection App is available here and the code for SVM modelling in MATLAB here.

## 4  Results

### 4.1  Computer Vision

So now the app can detect potholes at both day and night time and calculate its intrinsic parameters like length, width, area, the effect of riding both in the captured image and live stream then we tested the pothole detection app in real-time and it gave us impressive results. Figure 6 shows the detection during day time, the pothole's area is nearly 18% of the image (1040 x 780 pixels), and the number of potholes detected is one. As shown in Figure 7, the image was taken at night time, the pothole's area is around 32% of the image (1040 x 780 pixels), and the number of potholes detected is one.

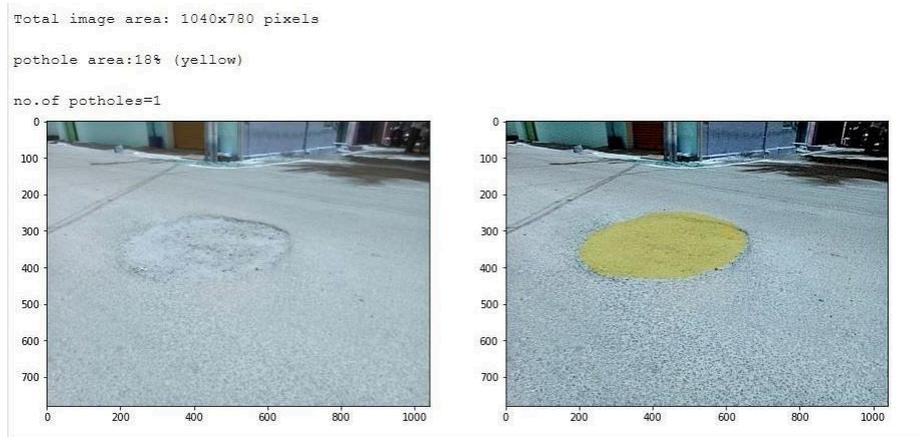

**Fig. 6.** Day time detection

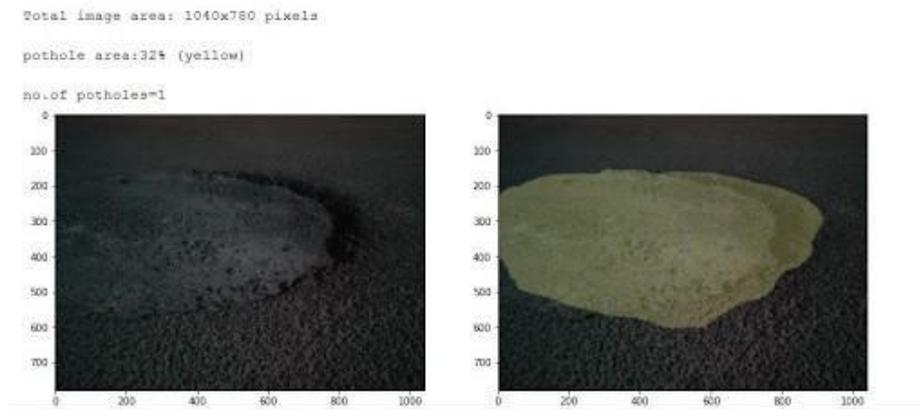

**Fig. 7.** Night time detection

### 4.2  Accelerometer and Gyroscope

The confusion matrix was generated, as shown in Figure 8, for a set of 26 data points obtained while driving on potholes and 26 data points while driving on a non-pothole area. The failure rate of 1.9% was present in the system, 1 time out of 102 times, a pothole road was classified as a plain road.

**Fig. 8.** Confusion matrix

This is reasonable for detection without any visual feed. Legend: Class 1 is Plain road and Class 0 is Pothole.

## 5  Future work

Integration of Computer vision and the Accelerometer prediction can be fused with Kalman filter. Idea and Codes can then be easily exported to embedded platforms for dedicated detection mechanism. SVM model can be improved by adding more features such as velocity, min-max of sensor values, classify road as good or bad.

## Acknowledgment

The authors would like to express their sincere thanks and deep sense of gratitude to their faculties, Dr. R. Menaka and Dr. S. Muthulakshmi, who organized this competition provided us with the opportunity to present this idea for the Ericssion Designathon competition. The Ericssion team also gave their insights and valuable feedback for us throughout the course of the project work.